\documentclass[11pt]{article}
\usepackage[final]{acl}

\usepackage{times}
\usepackage{latexsym}
\usepackage{amsmath, amsfonts, amssymb}
\usepackage{booktabs}
\usepackage{multirow}
\usepackage{xcolor, soul}
\usepackage{enumitem}
\usepackage{graphicx}
\usepackage{subcaption}
\usepackage{siunitx}
\usepackage[table]{xcolor}
\usepackage{caption}
\usepackage[T1]{fontenc}
\usepackage[utf8]{inputenc}
\usepackage{microtype}
\usepackage{inconsolata}

\title{Text-Attributed Knowledge Graph Enrichment with \\ Large Language Models for Medical Concept Representation}

\author{
Mohsen Nayebi Kerdabadi, Arya Hadizadeh Moghaddam, \\
\textbf{Chen Chen, Dongjie Wang, Zijun Yao}\thanks{Corresponding author.} \\
University of Kansas, USA \\
\texttt{\{mohsen.nayebi, a.hadizadehm, chenchen, wangdongjie, zyao\}@ku.edu}
}

\begin{document}
\maketitle

\newcommand{\model}{\textsc{MedCo}}

\begin{abstract}
In electronic health record (EHR) mining, learning high-quality representations of medical concepts (e.g., standardized diagnosis, medication, and procedure codes) is fundamental for downstream clinical prediction. However, robust concept representation learning is hindered by two key challenges: (i) clinically important cross-type dependencies (e.g., diagnosis-medication and medication-procedure relations) are often missing or incomplete in existing ontology resources, limiting the ability to model complex EHR patterns; and (ii) rich clinical semantics are often missing from structured resources, and even when available as text, are difficult to integrate with KG structure for representation learning. To address these challenges, we present \model{}, an LLM-empowered graph learning framework for medical concept representation. \model{} first builds a global knowledge graph (KG) over medical codes by combining statistically reliable associations mined from EHRs with type-constrained LLM prompting to infer semantic relations. It then utilizes LLMs to enrich the KG into a text-attributed graph by generating node descriptions and edge rationales, providing semantic signals for both concepts and their relationships. Finally, \model{} jointly trains a LoRA-tuned LLaMA text encoder with a heterogeneous GNN, fusing text semantics and graph structure into unified concept embeddings. Extensive experiments on MIMIC-III and MIMIC-IV show that \model{} consistently improves prediction performance and serves as an effective plug-in concept encoder for standard EHR pipelines.
\end{abstract}

\section{Introduction}

Electronic health records (EHRs) encode a patient's medical history as a high-dimensional and sparse sequence of diagnosis (dx), medication (rx), and procedure (px) concepts, which enable a broad range of clinical prediction tasks~\cite{choi2016retain, nayebi2023contrastive, moghaddam2024contrastive, nayebi2025multi}. A cornerstone in EHR mining is to learn medical concept representations that can capture the meaningful but complex dependency patterns (e.g., dx-rx treatment/contraindication, dx-px indication/care pathways, or dx-dx comorbidity) to benefit downstream outcome prediction. Therefore, utilizing knowledge graphs to model heterogeneous concepts and their relations becomes an appealing direction to improve concept representation learning.

However, the KGs most commonly used in practice are largely derived from existing medical ontologies, which either encode within-type parent-child hierarchies (e.g., ICD) or provide limited cross-type semantics (e.g., UMLS)\footnote{A broader comparison of existing medical ontologies is provided in Appendix~\ref{app:ontologies}, Table~\ref{tab:ontology_limits}.}. Although large language models (LLMs) have recently emerged as a promising tool for enriching KGs by proposing missing relations and generating structured semantic descriptions, unconstrained prompting can produce plausible but unsupported edges and inconsistent outputs for the same concepts. Furthermore, because many clinically meaningful relations are context-dependent and vary across cohorts, care settings, and time, LLM-based KG enrichment must validate proposed relations against what is actually observed in the target EHR dataset, ensuring that edges are not only clinically interpretable but also empirically supported. These requirements create a key tension: although LLMs encode broad biomedical knowledge, KG induction for clinical modeling must remain evidence-grounded, type-aware, and globally consistent.

Beyond \emph{which} relations a KG should contain, a second challenge is \emph{how} to incorporate node and edge information into concept embedding. Since each medical concept can be associated with rich LLM-generated semantics, such as indications, mechanisms of action, roles in care, and contextual constraints, it is natural to model the KG as a text-attributed graph in which nodes and edges are paired with descriptive text. This representation has the potential to combine the relational expressiveness of KGs with the semantic richness of LLM-generated knowledge. However, effectively learning from an LLM-enriched text-attributed graph is nontrivial. Graph neural networks (GNNs) excel at aggregating graph structure but are not designed to interpret long-form medical language. In contrast, LLMs can encode nuanced textual semantics but do not explicitly enforce or exploit global relational constraints. This challenge motivates a co-learning mechanism in which (i) an LLM encodes node- and edge-level textual context, while (ii) a relation-aware GNN propagates and refines these representations over the KG structure. In this way, textual semantics and graph structure can be jointly captured and aligned under task supervision.

In this work, we propose \model{}, which addresses the above challenges by constructing a clinically interpretable and empirically supported KG and learning text-attributed concept embeddings through a KG-LLM co-learning framework. Specifically, we first build a global heterogeneous KG over diagnosis, medication, and procedure codes by combining (i) statistically reliable associations from EHR data, such as intra-visit co-occurrences and cross-visit temporal transitions, with (ii) structured, type-constrained LLM prompting that assigns directed relation types and calibrated confidence scores. Using the resulting text-attributed KG, enriched with node descriptions and edge rationales, we then jointly train a LoRA-tuned LLaMA text encoder and a heterogeneous GNN to fuse text-derived clinical semantics with graph-derived relational structure into unified concept embeddings for downstream EHR prediction. Our contributions are as follows:
\begin{itemize}[leftmargin=*]
    \item We introduce an LLM-assisted pipeline to construct a global heterogeneous KG by combining EHR-derived co-occurrences and temporal statistics with type-constrained relation inference, yielding clinically meaningful relations among diagnosis, medication, and procedure codes.
    \item We enrich the KG into a text-attributed graph with LLM-generated node and edge semantics, including textual rationales, relation labels, confidence scores, and EHR-derived statistics.
    \item We propose \model{}, a KG-LLM co-learning framework that jointly fine-tunes a LLaMA text encoder and a heterogeneous GNN to learn unified medical concept embeddings.
    \item We show that \model{} serves as an effective plug-in concept encoder for standard EHR backbones through extensive experiments on MIMIC-III and MIMIC-IV for sequential diagnosis prediction.
\end{itemize}

\section{Methodology}

\subsection{Notation and Problem Definition}
\textbf{EHRs.}
We represent the EHR record for patient $j\in\mathcal{J}$ as a sequence of $T_j$ clinical visits, denoted by
$\mathcal{X}_j=\{V_{j,t}\}_{t=1}^{T_j}$.
Each visit $V_{j,t}$ is a set of $N_{j,t}$ medical codes,
$V_{j,t}=\{c_i\}_{i=1}^{N_{j,t}}$, where each code $c_i$ corresponds to a diagnosis (\textit{dx}), a medication (\textit{rx}), or a procedure (\textit{px}).
For brevity, we omit the patient index $j$ and visit index $t$.

\noindent\textbf{Healthcare Predictive Tasks.}  
Given a patient’s visit sequence \(\mathcal{X}_{j} = \{V_{1}, V_{2}, \ldots, V_{T_j}\}\), the objective is to predict clinical outcomes. These may involve binary tasks (e.g., mortality) or multi-label classification (e.g., diagnosis prediction). In this paper, we focus on predicting diagnosis codes for the next visit \(V_{T_j+1}\), a comprehensive task aimed at identifying potential diseases for future encounters.

\begin{figure*}[t]
    \centering
    \includegraphics[width=0.75\textwidth]{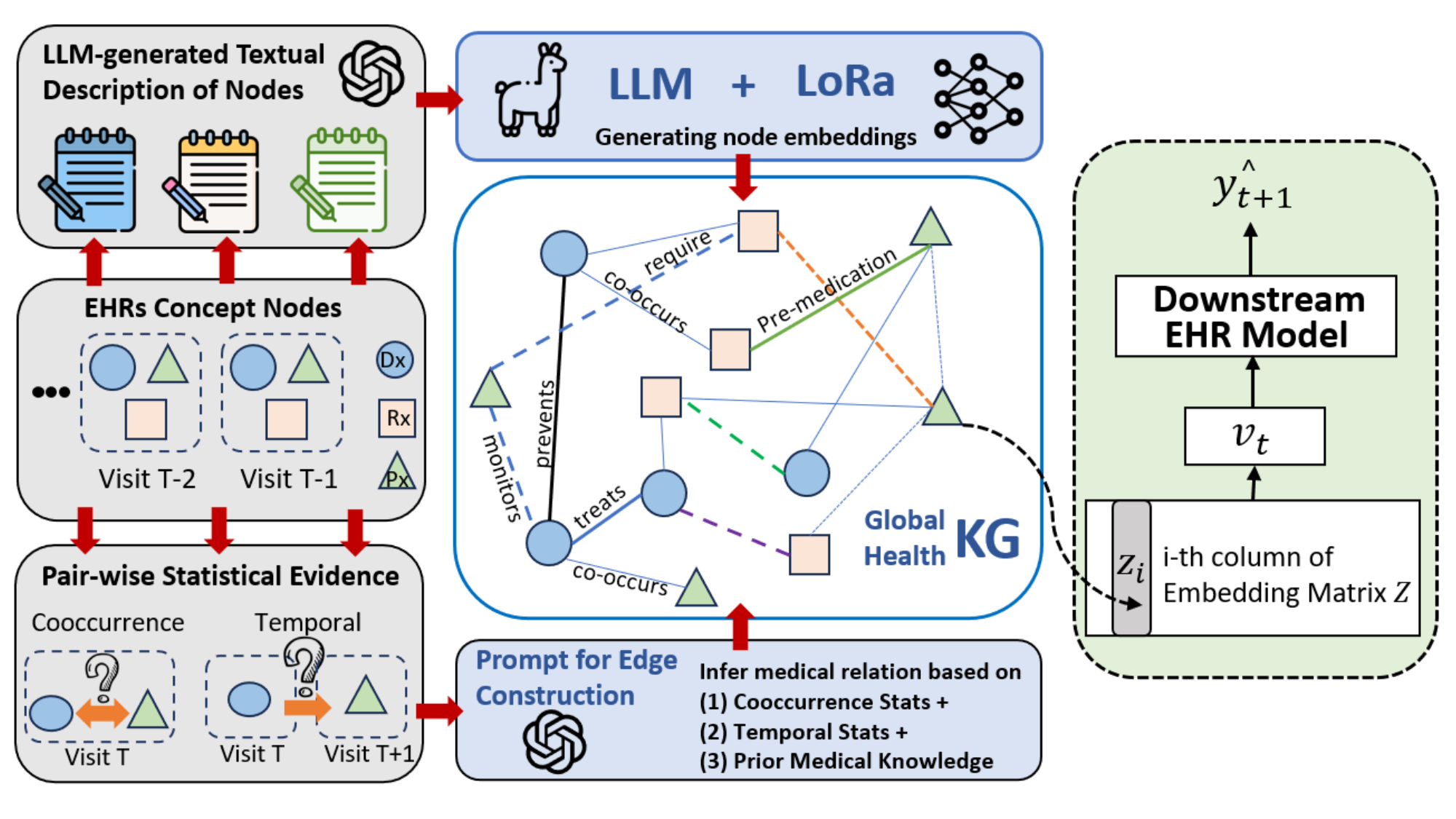} 
    \caption{Overview of \model{}. (1) Extract intra-visit co-occurrence and next-visit transition evidence from EHRs, and retain well-supported code pairs. (2) Use type-constrained, evidence-conditioned LLM prompting to assign directed relation types (with confidence and rationales), thereby constructing a heterogeneous KG. (3) Enrich nodes with LLM-generated concept descriptions and edges with relation metadata. (4) Jointly train a LoRA-tuned LLaMA encoder and a relation-aware GNN to learn medical concept embeddings for downstream prediction.}
    \label{fig:Model}
\end{figure*}

\subsection{Method Summary}
We present \model{}, which constructs an evidence-grounded, LLM-guided heterogeneous KG and learns plug-in medical concept embeddings through an LLM-GNN co-learning framework. As illustrated in Figure~\ref{fig:Model}, the framework consists of four steps: \textbf{(1) Evidence extraction.} Compute co-occurrence and next-visit transition statistics, and retain statistically significant code pairs. \textbf{(2) KG induction.} Use type-constrained, evidence-conditioned LLM prompting to assign directed relation types, confidence scores, and rationales, thereby constructing a heterogeneous KG. \textbf{(3) KG enrichment.} Enrich the KG with LLM-generated node descriptions and edge metadata, including relations, supporting evidence, confidence scores, and rationale embeddings. \textbf{(4) Co-learning.} Jointly train a LoRA-tuned LLaMA encoder and a relation-aware GNN to fuse textual semantics and graph structure into concept embeddings for downstream EHR prediction.

\subsection{LLM-Guided Construction of Global Health Knowledge Graph}
We construct a clinically grounded heterogeneous KG by inducing relations among diagnosis, procedure, and medication codes using longitudinal EHRs. The pipeline combines EHR-derived co-occurrence and temporal evidence with evidence-conditioned LLM prompting to infer directed relation types. It proceeds in two stages: (1) extract pair-level statistical evidence and (2) perform evidence-supported LLM relation inference.

\subsubsection{Pairwise Statistical Evidence Extraction from EHRs}
\label{sec:kg_construction}
We extract empirical evidence for candidate code pairs from EHRs in two steps: (i) computing intra-visit co-occurrence and next-visit transition statistics, and (ii) filtering to retain well-supported, clinically meaningful pairs. Each patient is represented as a sequence of visits, where each visit is a de-duplicated set of diagnosis (dx), procedure (px), and medication (rx) codes. These visit-level code sets are used to compute the corresponding marginal frequencies and transition statistics.

\paragraph{Pair-level evidence.}
For a directed pair $(c_i \!\rightarrow\! c_j)$, we compute three signals: a smoothed conditional probability, pointwise mutual information (PMI), and a chi-square test of dependence. We use a common formulation for both intra-visit co-occurrence and next-visit transitions. Let $x(\cdot)$ denote the appropriate count function: for co-occurrence, $x(c_i)$ is the number of visits containing $c_i$, and $x(c_i,c_j)$ is the number of visits in which $c_i$ and $c_j$ co-occur; for transitions, $x(c_i)$ is the number of source-visit occurrences of $c_i$, and $x(c_i,c_j)$ is the number of observed transitions from $c_i$ in $V_t$ to $c_j$ in $V_{t+1}$. Let $\mathcal{C}$ denote the set of distinct codes and $|\mathcal{C}|$ its cardinality.

\textbf{(1) Smoothed conditional probability.}
\[
P(c_j \mid c_i) \;=\; \frac{x(c_i,c_j)+\alpha}{x(c_i)+\alpha|\mathcal{C}|},
\]
where $\alpha$ is a Laplace smoothing constant used to mitigate sparsity and avoid zero probabilities.

\textbf{(2) PMI-style association.}
\[
\mathrm{PMI}(c_i,c_j) \;=\; \log_2\!\left(\frac{p(c_i,c_j)}{p_{\text{src}}(c_i)\,p_{\text{tgt}}(c_j)}\right),
\]
where $p(\cdot)$ denotes empirical probabilities estimated from counts, and $p_{\text{src}}$ and $p_{\text{tgt}}$ are the corresponding marginal probabilities. For intra-visit co-occurrence, $p_{\text{src}} = p_{\text{tgt}}$.

\textbf{(3) Statistical significance.}
We compute a chi-square $p$-value from a $2{\times}2$ contingency table to test dependence between $c_i$ and $c_j$ for both intra-visit co-occurrence and next-visit transition settings.

\paragraph{Evidence Filtering.}
For each unordered pair $(c_i,c_j)$, we compute intra-visit and temporal evidence (support counts, conditional probabilities, PMI, and $\chi^2$ $p$-values) and consolidate them into a unified table covering both directions $(c_i\!\rightarrow\!c_j)$ and $(c_j\!\rightarrow\!c_i)$. We discard pairs with low support, weak association (low conditional probability/PMI), or non-significant dependence (e.g., $p>0.05$), yielding a set of statistically reliable candidates. These pairs are then passed to the LLM for relation-type assignment to construct a clinically meaningful, evidence-grounded KG. A detailed breakdown is provided in Appendix~\ref{app:pair_stats}.

\subsubsection{Evidence-Supported LLM Prompting for Medical Relationship Induction}

Given the final set of code pair candidates, the last stage of our pipeline assigns each pair a clinically meaningful semantic relationship using an LLM. We design a structured, evidence-supported prompting framework that guides the LLM to choose the most appropriate relation between two medical concepts from a predefined, type-specific relation pool, explicitly grounding its decisions in clinical knowledge and EHR-derived evidence.

\paragraph{Type-Constrained Relationship Pool.}
Each code-type combination (dx--dx, rx--dx, px--dx, rx--rx, px--px, px--rx) is associated with a curated set of candidate relations grounded in clinical practice (Appendix~\ref{app:ontology}). The resulting relation inventory spans key semantic axes, including causality (e.g., \textit{causes}), temporal progression (e.g., \textit{leads\_to}), therapeutic intent (e.g., \textit{treats}, \textit{combination\_therapy\_with}), diagnostic usage (e.g., \textit{diagnostic\_of}), workflow structure (e.g., \textit{prerequisite\_for}), and safety considerations (e.g., \textit{contraindicated\_for}). When neither clinical priors nor statistical evidence supports a confident assignment, the relation inventory provides conservative abstention labels (\textit{no\_significant\_relation}, \textit{cannot\_decide}). Collectively, these relations define a concise, interpretable semantic space for modeling clinically meaningful dependencies in EHRs.

\begin{figure}[t]
    \hspace*{-0.25cm}
    \includegraphics[width=0.48\textwidth]{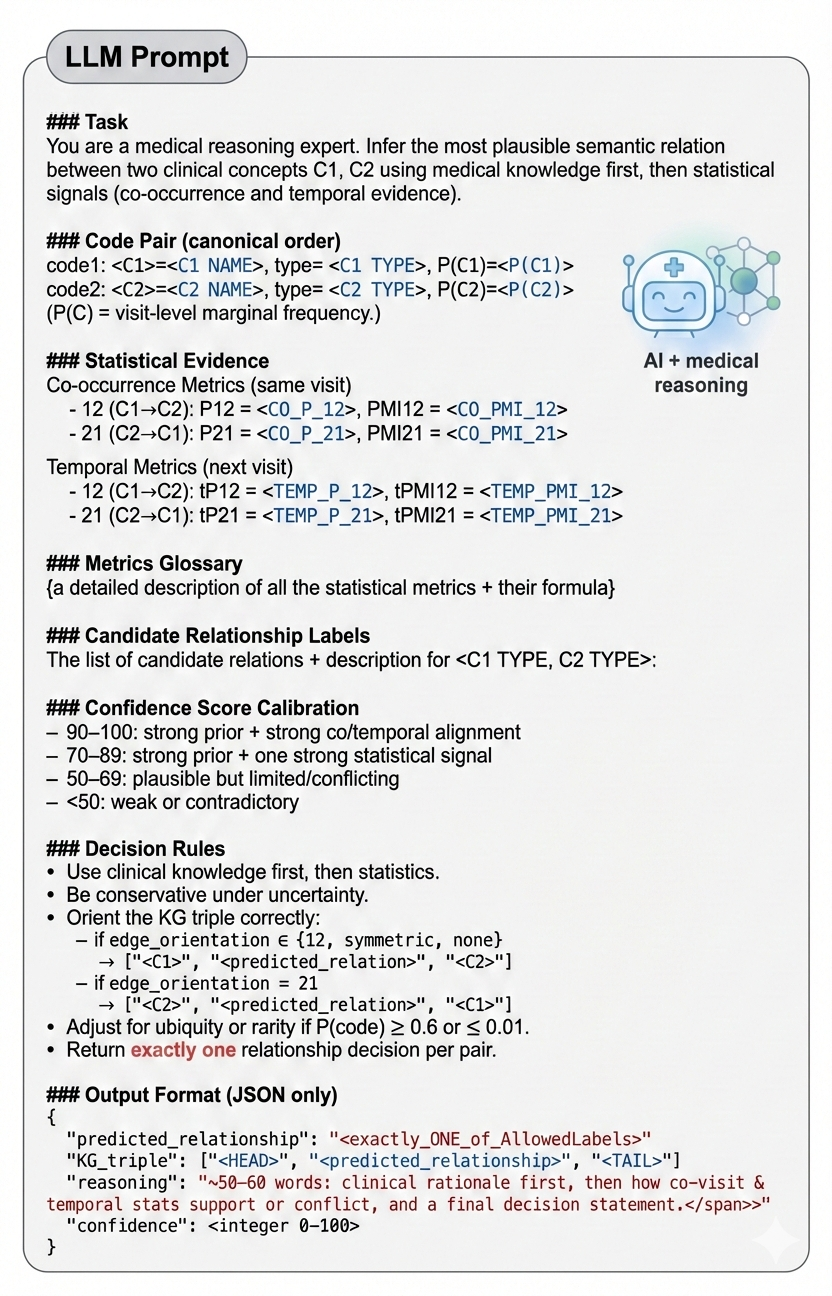}
    \caption{Prompt for Medical Relationship Induction.}
    \label{fig:prompt}
    \vspace{-5mm}
\end{figure}

\paragraph{Structured Prompting.}
For each code pair, we construct a prompt (Figure~\ref{fig:prompt}) that includes: (i) the code identifiers, textual names, parent categories, and marginal frequencies for both codes; (ii) directional statistical evidence from intra-visit co-occurrence and next-visit transitions (8 metrics) together with a brief metrics glossary; (iii) candidate relations permitted by the corresponding code-type combination (e.g., dx-rx); and (iv) explicit decision rubrics and confidence-calibration guidelines. The LLM is instructed to prioritize clinical plausibility, using statistical signals as supporting evidence, and to return a strict JSON object containing: (1) a single relation label, (2) an oriented KG triple, (3) a calibrated confidence score, and (4) a 50–60 word reasoning explaining the clinical rationale and how the co-occurrence and temporal statistics support or challenge the decision. This inference stage transforms medical knowledge encoded in the LLM, guided by statistically validated evidence, into a structured, relation-typed, and clinically interpretable knowledge graph that unifies EHR-derived evidence with clinical reasoning.

\paragraph{Clinical expert audit of induced edges.}
To assess potential hallucination in LLM-inferred relations, we conducted a targeted clinical expert audit of the constructed KG. We sampled 50 edges by selecting 5 edges from each of the 10 most frequent predicted relation types. Two independent frontline clinicians from the University of Kansas Health System reviewed each edge under general clinical knowledge and rated its correctness on a 1--5 scale. The mean score, averaged across the two reviewers for each edge, was $4.84 \pm 0.29$, indicating high agreement with the clinical validity of the sampled relations. Full protocol details and the relation-wise summary are provided in Appendix~\ref{app:expert_audit}.

\subsection{LLM-Based Contextual KG Enrichment}

Starting from the constructed medical KG, we further enrich nodes and edges with contextual information derived from a large language model (LLM). The LLM is treated as a high-coverage medical knowledge base: it provides clinically detailed descriptions for individual medical concepts and augments edge semantics using relation reasoning and EHR-derived evidence. The resulting features supply a semantic and knowledge-grounded context for downstream graph learning.

\paragraph{Node-Level Enrichment.}
For each medical code (diagnosis, procedure, or medication), we use a type-specific prompt to generate a concise, clinically focused description using an LLM (e.g., typical presentation, indications, role in care, and key nuances; see Appendix~\ref{app:node_prompt}). We attach this description to the corresponding KG node as textual attributes, which are subsequently encoded by \model{} during joint KG-LLM co-learning.

\paragraph{Edge-Level Enrichment.}
For each candidate code pair, the previous subsection yields (i) an oriented relation label, (ii) an LLM-generated confidence score and free-text rationale, and (iii) eight EHR-derived statistical metrics (co-occurrence and temporal transition statistics). We combine these signals to construct informative edge features. 

\subsection{\model{}: Co-Learning of LLM and GNN for Medical Concept Representations}
\label{sec:colearning}

Given the constructed KG, we learn unified node representations with \model{}, a co-learning medical concept representation framework that jointly fine-tunes a large language model (LLM) and a heterogeneous graph neural network (GNN). The LLM encodes each node’s textual description into an initial embedding, which is then passed to the GNN for neighborhood message passing over the KG. We train both components end-to-end; for the LLM, we use LLaMA-1B and fine-tune it with low-rank adaptation (LoRA). This co-learning setup combines the LLM’s semantic knowledge with the GNN’s ability to incorporate relational structure and EHR-derived evidence.

\paragraph{Architecture.}
Let $G=(V,E)$ denote the global heterogeneous medical knowledge graph, where $V$ is the set of medical concepts and $E$ is the set of directed edges. Each node $i\in V$ represents a medical concept with node type $\tau(i)\in\mathcal{T}$, where $\mathcal{T}$ denotes the set of node types (diagnosis, medication, and procedure), and is associated with a textual description $s_i$ (Appendix~\ref{app:node_prompt}). Each directed edge $(j\!\to\! i)\in E$ is associated with a relation type $r_{ji}\in\mathcal{R}$, where $\mathcal{R}$ is the set of relation types, and an edge feature vector $\mathbf{e}_{ji}\in\mathbb{R}^{d_e}$.

A LLaMA-based text encoder $f_\theta$, fine-tuned with LoRA adapters, maps the description $s_i$ to a text embedding, which is then projected into the GNN hidden space using a type-specific linear map $\mathbf{W}_{\tau(i)}\in\mathbb{R}^{d\times d_L}$:
\begin{equation}
\begin{aligned}
\mathbf{z}_i^{\text{text}} &= f_\theta(s_i)\in\mathbb{R}^{d_L},\\
\mathbf{h}_i^{(0)} &= \mathbf{W}_{\tau(i)}\,\mathbf{z}_i^{\text{text}}\in\mathbb{R}^{d}.
\end{aligned}
\end{equation}

These embeddings initialize a relation-aware heterogeneous GNN $g_\phi$ that incorporates both relation types and edge features during message passing. Let $\mathcal{N}(i)=\{j\in V:(j\!\to\! i)\in E\}$ denote the set of in-neighbors of node $i$. For layer $\ell\in\{0,\dots,L-1\}$, node states are updated as
\begin{equation}
\label{eq:gnn_update}
\begin{aligned}
\mathbf{u}_{ji}^{(\ell)}
&= \mathrm{MSG}^{(\ell)}\!\left(
\mathbf{h}_i^{(\ell)}, \mathbf{h}_j^{(\ell)}, \mathbf{e}_{ji}, r_{ji}
\right),\\
\mathbf{m}_i^{(\ell)}
&= \mathrm{AGG}^{(\ell)}\!\left(
\left\{
\mathbf{u}_{ji}^{(\ell)} : j \in \mathcal{N}(i)
\right\}
\right),\\
\mathbf{h}_i^{(\ell+1)}
&= \mathrm{UPDATE}^{(\ell)}\!\left(
\mathbf{h}_i^{(\ell)}, \mathbf{m}_i^{(\ell)}
\right).
\end{aligned}
\end{equation}
Here, $\mathrm{MSG}^{(\ell)}(\cdot)$ denotes the message function that computes the edge-specific message $\mathbf{u}_{ji}^{(\ell)}$ from node $j$ to node $i$; $\mathrm{AGG}^{(\ell)}(\cdot)$ denotes a permutation-invariant aggregation operator over incoming messages; and $\mathrm{UPDATE}^{(\ell)}(\cdot)$ denotes the node-state update function at layer $\ell$. Intuitively, the LLM provides semantic information from textual concept descriptions, while the heterogeneous GNN incorporates relational structure and EHR-derived evidence. Joint end-to-end training fuses these two sources into clinically meaningful concept representations.

\paragraph{LLM-GNN Training Strategy.}
We jointly train the LLM-based node encoder and the GNN, but a practical challenge arises during mini-batch training. Each batch contains multiple patients, and the union of diagnosis, medication, and procedure codes appearing across their visit sequences can be large. If we update the LLM representations for all such codes at every iteration, the cost becomes prohibitively expensive, especially given the size of the medical code vocabulary. A naive workaround is to update only a random subset of batch-activated codes per iteration, but this has two drawbacks: (i) rare yet clinically important codes may receive very few or no updates, and (ii) highly frequent codes may be updated disproportionately often, leading to unbalanced adaptation of the LLM.

To address this issue, we maintain lightweight per-code statistics that track how often each code appears in the training data and how many times its LLM representation has been updated. We then adopt a two-phase sampling schedule for LoRA updates. In the early phase, we prioritize codes with the fewest updates (“least-updated-first”), encouraging broad coverage and ensuring that all observed codes receive at least a few LLM updates. In the later phase, each iteration mixes a subset of the least-updated codes with a subset of the most frequently occurring codes from the current batch, allowing the model to further refine high-impact nodes without neglecting the long tail. This strategy amortizes LLM computation while enabling controlled, coverage-aware fine-tuning of the LLM representations used by the GNN.

\subsection{Integrating \model{} into EHR Models}

 As a medical concept encoder, \model{} can be integrated into standard EHR models and trained end-to-end; it improves concept representation learning and consequently downstream predictive performance. Given the learned concept (code) embeddings $\mathbf{z}_{1}, \mathbf{z}_{2}, \dots, \mathbf{z}_{N}$, we form the embedding matrix $\mathbf{Z}\in\mathbb{R}^{d\times N}$, where $\mathbf{z}_{i}$ is the $i$-th column of $\mathbf{Z}$. This matrix is then used by a downstream task model. Considering sequential diagnosis prediction, which maps a sequence of visits to the diagnoses of the next visit:
$f:\{V_{1},V_{2},\ldots,V_{t}\}\rightarrow \mathbf{\hat{y}}_{t+1}$, where $\mathbf{\hat{y}}_{t+1}\in\mathbb{R}^{N_{dx}}$ is a multi-hot vector and $N_{dx}$ denotes the total number of diagnosis codes:
\begin{equation}
\label{eq:downstream_task}
\begin{aligned}
    \mathbf{Z} = [\mathbf{z}_{1}, \mathbf{z}_{2}, \ldots, \mathbf{z}_{N}] &\gets \text{\model{}}(\mathbf{x}_{1},\mathbf{x}_{2}, \ldots, \mathbf{x}_{N}) \\
    \mathbf{v}_{1}, \mathbf{v}_{2}, \ldots, \mathbf{v}_{t} &= \mathbf{Z}[\mathbf{u}_{1}, \mathbf{u}_{2}, \ldots, \mathbf{u}_{t}] \\
    \mathbf{h}_{p} &= \text{Model}(\mathbf{v}_{1}, \mathbf{v}_{2}, \ldots, \mathbf{v}_{t}) \\
    \mathbf{\hat{y}}_{t+1} &= \text{Sigmoid}(\mathbf{W} \mathbf{h}_{p} + \mathbf{b})
\end{aligned}
\end{equation}
For each visit $V_t$, we compute $\mathbf{v}_t=\mathbf{Z}\mathbf{u}_t$, where $\mathbf{u}_t\in\{0,1\}^{N}$ is a multi-hot code indicator and $\mathbf{Z}$ is the learned concept embedding matrix. The sequence $\{\mathbf{v}_1,\ldots,\mathbf{v}_t\}$ is encoded by a backbone model to produce a patient representation $\mathbf{h}_p$. The final prediction is obtained by a linear classification layer followed by a sigmoid activation. We train with multi-label cross-entropy between $\hat{\mathbf{y}}_{t+1}$ and $\mathbf{y}_{t+1}$ at each timestep.

\section{Experimental Setting}

\begin{figure*}[t]
    \setlength{\abovecaptionskip}{3pt} 
    \begin{center}
    \includegraphics[width=1.0\linewidth]{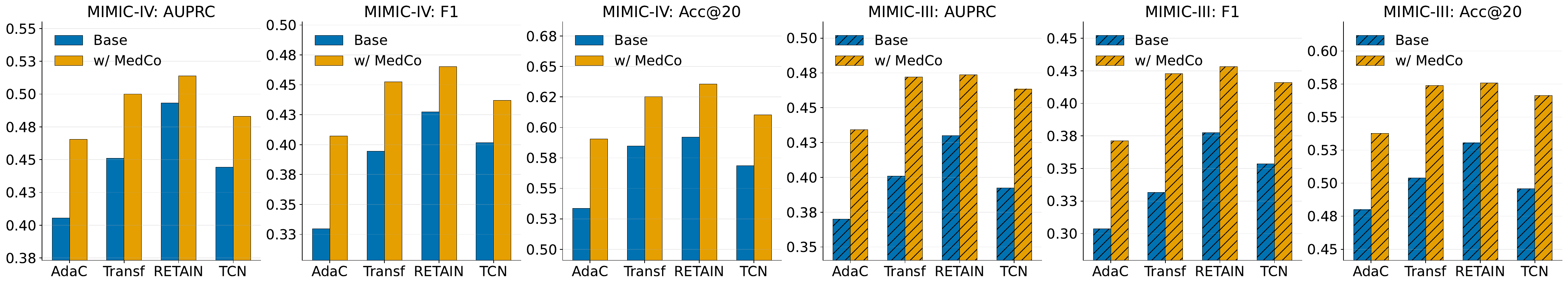}
    \caption{Performance gains from integrating \model{} into four diagnosis prediction backbones (plug-in analysis) on MIMIC-III and MIMIC-IV.} 
    \label{fig:enhancement}
    \end{center}
    \vspace{-3mm}
\end{figure*}

\begin{table*}[t]
    \centering
    \scriptsize
    \setlength{\tabcolsep}{5pt}
    \renewcommand{\arraystretch}{1.05}
    \caption{Performance comparison on MIMIC-III and MIMIC-IV. We report overall AUPRC, F1, and Acc@k, and stratified AUPRC by diagnosis label frequency quartiles (0--25\%, 25--50\%, 50--75\%, 75--100\%). The first row corresponds to the base Transformer, and subsequent rows denote plug-in variants (e.g., GRAM = Base + GRAM).}
    \vspace{-2mm}
    \begin{tabular}{c l ccccc cccc}
        \toprule
        & \multirow{2}{*}{\textbf{Model}} &
        \multicolumn{5}{c}{\textbf{General Performance}} &
        \multicolumn{4}{c}{\textbf{Label Category Performance (AUPRC)}} \\
        \cmidrule(lr){3-7} \cmidrule(lr){8-11}
        & & \textbf{AUPRC} & \textbf{F1} & \textbf{Acc@15} & \textbf{Acc@20} & \textbf{Acc@30}
          & \textbf{0--25\%} & \textbf{25--50\%} & \textbf{50--75\%} & \textbf{75--100\%} \\
        \midrule

        \multirow{10}{*}{\rotatebox{90}{\textbf{MIMIC-III}}}
        & Base  & 41.00 & 33.16 & 47.20 & 50.40 & 58.80 & 40.60 & 47.30 & 72.80 & 78.40 \\
        & GRAM  & 41.70 & 34.60 & 48.60 & 51.90 & 59.80 & 42.20 & 50.10 & 74.10 & 79.30 \\
        & MMORE & 42.60 & 35.30 & 49.20 & 52.60 & 60.40 & 42.80 & 51.20 & 74.80 & 80.00 \\
        & KAME  & 42.18 & 35.07 & 48.97 & 52.06 & 60.06 & 41.84 & 49.76 & 74.06 & 79.14 \\
        & G-BERT& 42.47 & 35.38 & 49.26 & 52.33 & 60.27 & 42.06 & 50.04 & 74.24 & 79.28 \\
        & HAP   & 42.36 & 35.17 & 49.11 & 52.18 & 60.18 & 41.95 & 49.92 & 74.13 & 79.24 \\
        & ADORE & 42.58 & 35.43 & 49.33 & 52.47 & 60.32 & 42.18 & 50.09 & 74.29 & 79.31 \\
        & KAMPNet & 43.06 & 35.96 & 49.88 & 53.07 & 60.86 & 42.63 & 50.83 & 74.77 & 79.89 \\
        & GraphCare & 43.35 & 35.46 & 52.76 & 56.00 & 62.75 & 44.80 & 58.16 & 70.97 & 65.72 \\
        & LINKO & 44.91 & 38.20 & 52.30 & 55.62 & 62.80 & 44.80 & 55.20 & \textbf{76.43} & 83.41 \\
        \rowcolor{gray!10}
        & \textbf{\model{}} & \textbf{47.21} & \textbf{42.28} & \textbf{54.20} & \textbf{57.40} & \textbf{64.39}
        & \textbf{47.67} & \textbf{66.29} & 76.02 & \textbf{86.70} \\
        \addlinespace[1mm]
        
        \multirow{10}{*}{\rotatebox{90}{\textbf{MIMIC-IV}}}
        & Base  & 45.10 & 39.10 & 54.80 & 58.10 & 64.70 & 46.80 & 53.60 & 53.90 & 77.20 \\
        & GRAM  & 46.30 & 40.00 & 55.60 & 58.90 & 65.70 & 47.70 & 54.60 & 54.80 & 77.90 \\
        & MMORE & 46.90 & 40.60 & 56.10 & 59.40 & 66.10 & 48.20 & 55.10 & 55.70 & 78.40 \\
        & KAME  & 45.97 & 39.66 & 55.27 & 58.57 & 65.26 & 47.16 & 54.24 & 54.28 & 77.63 \\
        & G-BERT& 46.37 & 40.07 & 55.68 & 58.96 & 65.66 & 47.58 & 54.73 & 54.87 & 78.04 \\
        & HAP   & 46.33 & 39.88 & 55.57 & 58.92 & 65.58 & 47.52 & 54.57 & 54.83 & 77.92 \\
        & ADORE & 46.52 & 40.23 & 55.84 & 59.09 & 65.79 & 47.74 & 54.84 & 55.06 & 78.13 \\
        & KAMPNet & 47.08 & 40.96 & 56.46 & 59.77 & 66.47 & 48.26 & 55.57 & 56.16 & 78.68 \\
        & GraphCare & 46.90 & 40.12 & 55.89 & 59.14 & 65.71 & 48.31 & \textbf{60.31} & 73.70 & 66.33 \\
        & LINKO & 48.14 & 42.05 & 57.78 & 61.21 & 67.97 & 49.61 & 56.87 & 57.58 & 80.27 \\
        \rowcolor{gray!10}
        & \textbf{\model{}} & \textbf{50.00} & \textbf{45.25} & \textbf{59.22} & \textbf{62.54} & \textbf{68.71}
        & \textbf{52.01} & 57.70 & \textbf{59.35} & \textbf{81.19} \\
        \bottomrule
    \end{tabular}
    \vspace{-0.2cm}
    \label{tab:baseline_comparison_w_rare_codes}
\end{table*}

\textbf{Datasets.} We evaluate on two public EHR benchmarks, MIMIC-III \citep{johnson2016mimic} and MIMIC-IV \citep{johnson2023mimic}. We map ICD diagnosis/procedure codes to CCS categories and medications from National Drug Codes (NDC) to the Anatomical Therapeutic Chemical (ATC) classification. Table~\ref{tab:Data-Statistics} reports cohort statistics. The task is next-visit diagnosis prediction over imbalanced label spaces (515 codes in MIMIC-III; 562 in MIMIC-IV). We use \texttt{Llama-3.2-1B} \citep{grattafiori2024llama3,meta2024llama32_1b} as the text encoder and fine-tune it with LoRA using rank $r{=}8$ and $\alpha{=}32$. The heterogeneous GNN encoder has $2$ layers with $1$ attention head per layer and dropout $0.4$. The source code is publicly available.\footnote{\url{https://github.com/mohsen-nyb/MedCo.git}}

\begin{table}[t]
\centering
\caption{Data statistics for MIMIC-III and MIMIC-IV.}
\label{tab:Data-Statistics}
\vspace{-2mm}
\scriptsize
\setlength{\tabcolsep}{4pt}
\renewcommand{\arraystretch}{1.00}
\begin{tabular}{lcc}
\toprule
\textbf{Metric} & \textbf{MIMIC-III} & \textbf{MIMIC-IV} \\
\midrule
\# Patients                 & 7,515   & 18,829 \\
\# Visits (samples)         & 12,430  & 25,028 \\
\# Labels/sample            & 12.31   & 10.56  \\
\# Unique conditions (ICD)  & 515     & 562    \\
\# Conditions/sample        & 26.85   & 59.50  \\
\# Drugs/sample             & 70.10   & 118.16 \\
\# Unique drugs             & 471     & 510    \\
\# Procedures/sample        & 6.49    & 5.41   \\
\# Unique procedures        & 280     & 322    \\
\bottomrule
\end{tabular}
\vspace{-5mm}
\end{table}

\noindent\textbf{Evaluation Metrics.} We report mean results over 5 folds for \textbf{AUPRC} (area under the precision-recall curve; also stratified by label frequency to assess performance across code rarity levels), \textbf{Acc@k} (top-$k$ accuracy normalized by $\min(k, |\mathbf{y}_{t+1}|)$), and \textbf{F1} (harmonic mean of precision and recall).

\section{Evaluation Results}
\vspace{-3mm}
To assess \model{}, we address the following research questions:
\textbf{RQ1:} Does \model{} improve downstream EHR prediction when used as a plug-in medical concept encoder?
\textbf{RQ2:} How does \model{} compare with existing medical code encoders?
\textbf{RQ3:} How do individual components of \model{} and different KG edge categories contribute to overall performance?
\textbf{RQ4:} How well does \model{} mitigate data scarcity and improve performance on rare conditions?

\begin{figure*}[t]
    \setlength{\abovecaptionskip}{2pt} 
    \begin{center}
    \includegraphics[width=0.9\linewidth]{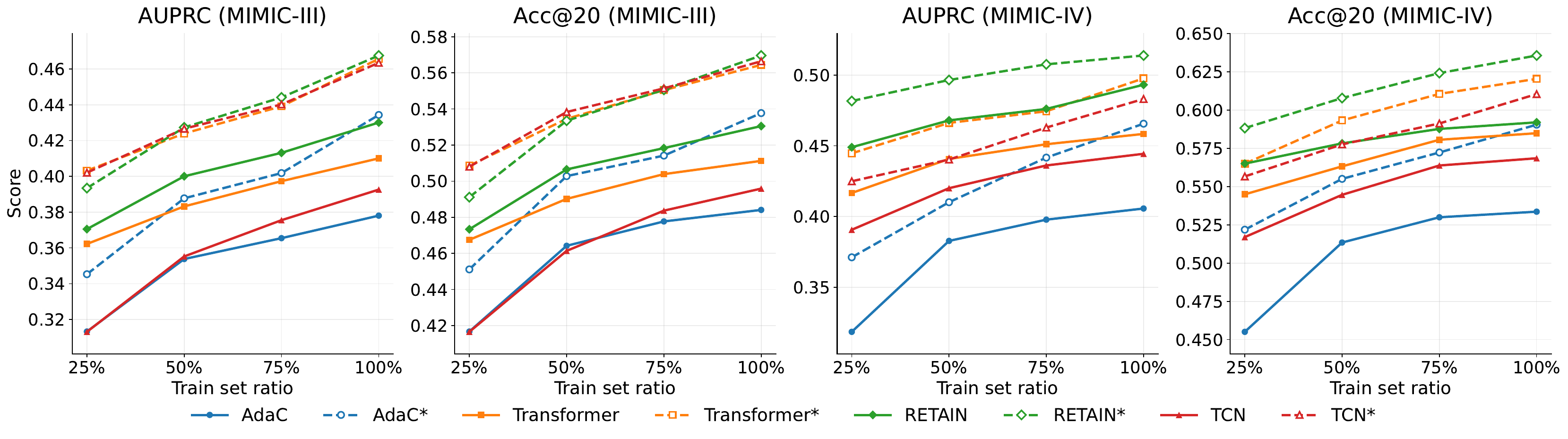}
    \caption{Performance evaluation across different training set sizes using the MIMIC-III and MIMIC-IV datasets. An asterisk (*) next to each encoder indicates the integration of \model{} to that model, e.g., Transformer* = Transformer + \model{}.}
    \label{fig:enhancement_rare_code}
    \end{center}
    \vspace{-5mm}
\end{figure*}

\begin{table}[t]
\centering
\scriptsize
\setlength{\tabcolsep}{3pt}
\renewcommand{\arraystretch}{1.12}
\caption{Component-wise ablation of \model{} on MIMIC-IV and MIMIC-III.}
\label{tab:component_ablation_rotated_pretty}
\vspace{-1mm}
\begin{tabular}{c l ccc}
\toprule
\textbf{} & \textbf{Model} & \textbf{PRAUC} & \textbf{F1} & \textbf{Acc@20} \\
\midrule
\multirow{5}{*}{\rotatebox{90}{\textbf{MIMIC-IV}}}
& Base (Transformer) & 45.10 & 39.10 & 58.10 \\
& Base + KG & 48.55 & 42.60 & 60.90 \\
& Base + KG + Edge feat. & 48.72 & 42.85 & 61.10 \\
& Base + KG + Edge feat. + LLM (freeze) & 49.38 & 43.71 & 61.47 \\
& \cellcolor{gray!10}\textbf{Base + KG + Edge feat. + LLM (LoRA)}
  & \cellcolor{gray!10}\textbf{50.00}
  & \cellcolor{gray!10}\textbf{45.25}
  & \cellcolor{gray!10}\textbf{62.54} \\
\addlinespace[0.6mm]
\multirow{5}{*}{\rotatebox{90}{\textbf{MIMIC-III}}}
& Base (Transformer) & 41.00 & 33.16 & 50.40 \\
& Base + KG & 45.79 & 38.90 & 55.60 \\
& Base + KG + Edge feat. & 45.91 & 39.20 & 55.85 \\
& Base + KG + Edge feat. + LLM (freeze) & 46.10 & 40.45 & 56.44 \\
& \cellcolor{gray!10}\textbf{Base + KG + Edge feat. + LLM (LoRA)}
  & \cellcolor{gray!10}\textbf{47.21}
  & \cellcolor{gray!10}\textbf{42.28}
  & \cellcolor{gray!10}\textbf{57.40} \\
\bottomrule
\end{tabular}
\vspace{-5mm}
\end{table}

\subsection{RQ1: Plug-in Enhancement Evaluation}
We hypothesize that integrating \model{}, our medical concept encoder, into standard EHR backbones improves downstream performance by enhancing concept representations. To test this hypothesis, we incorporate \model{} into four representative models: (1) \textbf{AdaCare}~\cite{ma2020adacare}, an explainable architecture that models multi-scale biomarker variations; (2) \textbf{Transformer}~\citep{vaswani2017attention}, based on self-attention; (3) \textbf{RETAIN}~\citep{choi2016retain}, an RNN with reverse-time, two-level attention; and (4) \textbf{TCN}~\citep{bai2018empirical}, which uses causal convolutions for temporal modeling. We compare each backbone with and without \model{}. As shown in Figure~\ref{fig:enhancement}, \model{} consistently improves performance across all models, demonstrating its effectiveness as a plug-in concept encoder.

\subsection{RQ2: Baseline Comparison}
We compare \model{} against widely used medical concept encoders, including the \textbf{base Transformer}~\cite{vaswani2017attention} (i.e., no concept encoder), hierarchy-driven ontology methods such as \textbf{GRAM}~\citep{choi2017gram}, \textbf{MMORE}~\citep{song2019medical}, \textbf{KAME}~\citep{ma2018kame}, and \textbf{HAP}~\citep{zhang2020hierarchical}, as well as KG-based approaches that leverage external knowledge resources. Among these, \textbf{ADORE}~\citep{cheong2023adaptive} relies on SNOMED, while \textbf{GraphCare}~\cite{jiang2023graphcare} leverages UMLS-derived edges together with LLM-derived edges. \textbf{KAMPNet}~\citep{an2023kampnet} and \textbf{LINKO}~\cite{nayebi2025multi} construct cross-type connections but do not explicitly model rich semantic relation types. All encoders are integrated into the same downstream backbone for a controlled comparison. As shown in Table~\ref{tab:baseline_comparison_w_rare_codes} (General Performance), \model{} achieves the best overall performance, indicating that our evidence-grounded, heterogeneous, text-attributed KG provides richer and more task-relevant semantics than existing medical concept baselines.

Furthermore, to evaluate performance under data scarcity, we stratify diagnosis codes into four frequency quartiles (0–25\%, 25–50\%, 50–75\%, 75–100\%), where 0–25\% contains the rarest codes. This analysis probes whether \model{} can better propagate informative signals to low-frequency concepts. Table~\ref{tab:baseline_comparison_w_rare_codes} (Label Category Performance) shows that \model{} improves AUPRC across all bands, with pronounced gains on rare codes.

\begin{figure}[t]
    \centering \includegraphics[width=0.90\linewidth]{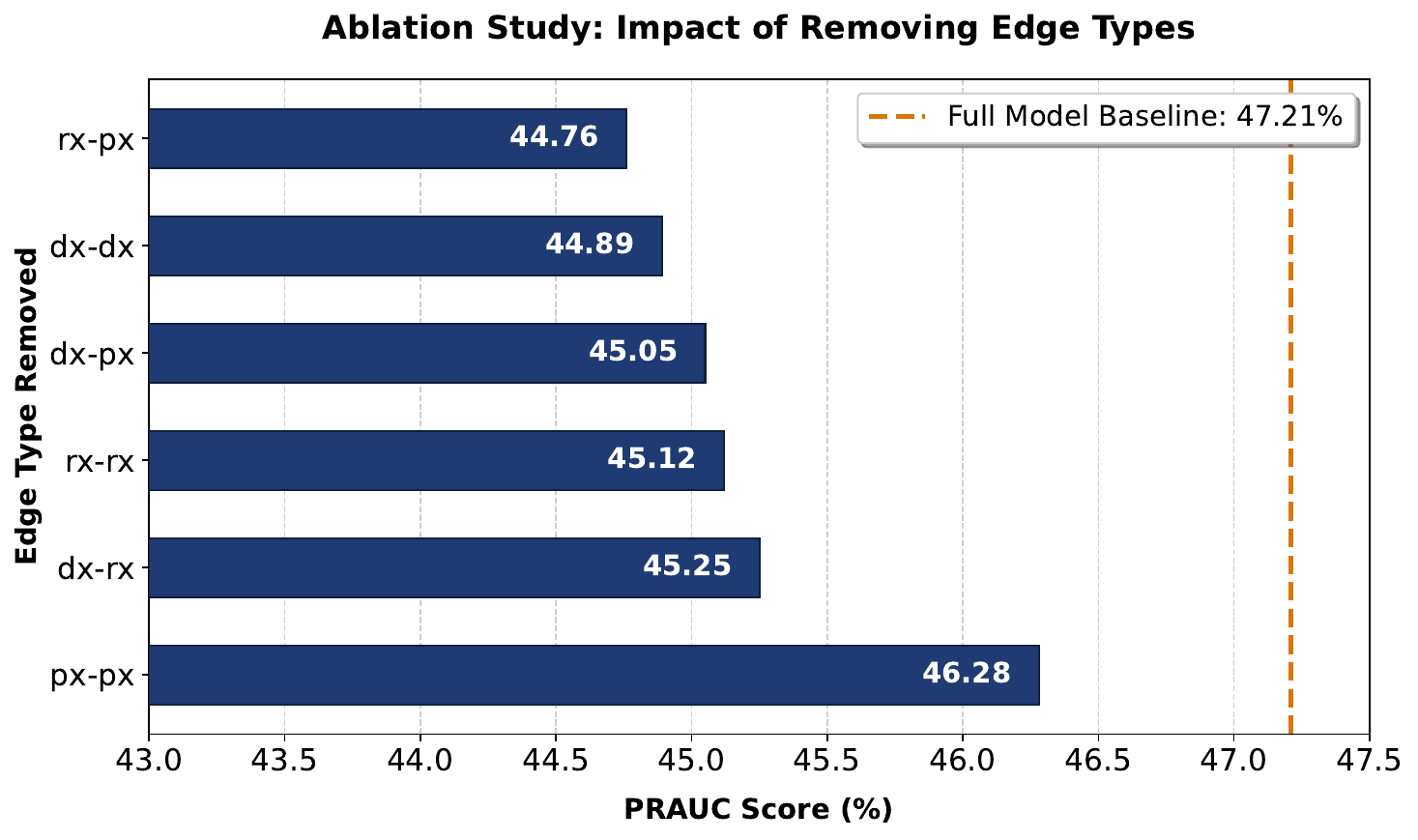}
    \caption{Ablation study showing the impact of removing each edge type on PRAUC for MIMIC-III. The dashed vertical line indicates the full-model baseline. }
    \label{fig:ablation_edge_type}
    \vspace{-0.4cm}
\end{figure}

\subsection{RQ3: Ablation Study}
Table~\ref{tab:component_ablation_rotated_pretty} shows progressive gains as components are added to \model{}. Starting from the base Transformer, adding the induced KG with a relation-aware GNN provides the largest boost, highlighting the value of explicit cross-domain structure. Edge features yield additional improvements by injecting LLM clinical reasoning and EHR-derived association evidence into message passing. Adding LLM-based node descriptions (frozen encoder) consistently boosts performance, indicating that clinical semantics from natural language complements graph signals. Finally, enabling LoRA fine-tuning delivers the best results, showing that adapting the LLM to the cohort and task produces more informative node representations and maximizes downstream accuracy. Figure~\ref{fig:ablation_edge_type} presents an edge-category ablation of the induced KG. For each bar, we remove all edges belonging to a given relation family (e.g., rx-px removes every medication-procedure edge) while keeping the rest of the graph unchanged, and report the resulting PRAUC. Larger drops indicate edge categories that contribute more to downstream performance.

\subsection{Sensitivity to the LLM Backbone}
We further evaluate whether \model{} is sensitive to the choice of LLM backbone for node-text encoding. While our main experiments use \texttt{Llama-3.2-1B}, the framework is model-agnostic and can be instantiated with other open-source LLMs. Under the same LoRA fine-tuning setup, we additionally test \texttt{Gemma-2-2B}~\citep{gemma2024gemma2,google2024gemma2_2b}, \texttt{Qwen2.5-1.5B}~\citep{hui2024qwen2,qwen2024qwen25_15b}, \texttt{Qwen2.5-0.5B}~\citep{hui2024qwen2,qwen2024qwen25_05b}, and \texttt{SmolLM2-1.7B}~\citep{allal2025smollm2,huggingfacetb2025smollm2_17b} on MIMIC-IV. As shown in Table~\ref{tab:llm_backbone_sensitivity}, performance remains consistently strong across backbones, indicating that \model{} is not tied to a specific LLM family. Larger backbones provide only modest gains, suggesting that downstream performance is driven primarily by the evidence-grounded KG construction and graph propagation components.

\begin{table}[t]
\centering
\small
\begin{tabular}{lcc}
\toprule
\textbf{LLM backbone} & \textbf{AUPRC} & \textbf{Acc@20} \\
\midrule
Llama-3.2-1B & 50.00 & 62.54 \\
Gemma-2-2B & 50.31 & 62.89 \\
Qwen2.5-1.5B & 50.16 & 62.67 \\
Qwen2.5-0.5B & 49.31 & 61.44 \\
SmolLM2-1.7B & 50.11 & 62.61 \\
\bottomrule
\end{tabular}
\caption{Sensitivity of \model{} to the choice of LLM backbone on MIMIC-IV.}
\label{tab:llm_backbone_sensitivity}
\vspace{-4mm}
\end{table}

\begin{table*}[ht]
\centering
\small
\setlength{\tabcolsep}{4pt}
\begin{tabular}{lcccccc}
\toprule
\textbf{Variant} & \textbf{AUPRC} & \textbf{Acc@20} & \textbf{Train(s)} & \textbf{Infer(s)} & \textbf{Peak train mem (MiB)} & \textbf{Peak infer mem (MiB)} \\
\midrule
Base (no KG, no LLM/GNN) & 45.10 & 58.10 & 22.0 & 1.30 & 169 & 106 \\
GNN only (KG, no LLM) & 48.72 & 61.10 & 42.7 & 3.19 & 1,273 & 518 \\
LLM frozen ($K{=}0$) + GNN & 49.38 & 61.47 & 40.0 & 3.19 & 1,273 & 518 \\
LLM LoRA + GNN ($K{=}5$) & 49.70 & 62.35 & 183.4 & 3.19 & 14,477 & 518 \\
LLM LoRA + GNN ($K{=}10$) & 50.00 & 62.54 & 275.4 & 3.19 & 24,260 & 518 \\
\bottomrule
\end{tabular}
\caption{Cost-accuracy trade-off on MIMIC-IV with batch size 128. Experiments were run on a machine with an Intel Xeon Silver 4214R CPU (24 cores), 256\,GiB RAM, and one RTX A6000 GPU (48\,GB). For compactness, \textit{Train(s)} denotes training time per epoch, and \textit{Infer(s)} denotes inference time per epoch for the test set.}
\label{tab:cost_accuracy_tradeoff}
\vspace{-4mm}
\end{table*}

\subsection{RQ4: Data Insufficiency Analysis}

To evaluate \model{} under data scarcity, we subsample the training set to simulate limited-data regimes. In the plug-in setting, Figure~\ref{fig:enhancement_rare_code} shows that \model{} delivers substantial gains for downstream EHR models even when training data are reduced, highlighting its robustness to data insufficiency.

\section{Cost-Accuracy Trade-off}
\vspace{-2mm}
Although jointly training the LLM and GNN yields the best performance, it can be computationally expensive. \model{} mitigates this cost in two ways: (i) LoRA fine-tunes only a small set of adapter parameters while freezing the backbone LLM, and (ii) node representations are cached and only a selected subset of codes is updated per epoch. At inference time, the LLM is removed from the loop, and prediction uses the cached concept representations together with the GNN and downstream EHR encoder, making deployment much lighter than training. Table~\ref{tab:cost_accuracy_tradeoff} summarizes the efficiency-accuracy trade-off on MIMIC-IV. The results show a clear frontier: KG-based variants substantially improve performance over the base model with moderate additional cost, while LoRA-based joint training achieves the highest accuracy at greater training cost. Inference remains lightweight and identical across KG-based variants because the LLM is not used at test time. This suggests that practitioners can prefer frozen-LLM or smaller-$K$ settings when efficiency matters most, and larger-$K$ LoRA training when accuracy is the priority.

\section{Related Work}
\label{related_work}
\vspace{-0.1cm}

The growing availability of EHR data has driven rapid progress in clinical prediction, evolving from early sequential architectures \citep{choi2016doctor} to attention-based methods \citep{choi2016retain, hu2025recurrent}, transformer models \citep{li2020behrt, nayebi2023contrastive}, and graph neural networks \citep{lu2021self, xu2022counterfactual, yang2023molerec, poulain2024graph}.

A line of work strengthens code embeddings by injecting hierarchical ontology structure into models. GRAM \citep{choi2017gram} represents each concept by mixing it with its ancestors, while MMORE \citep{song2019medical} extends this idea with multiple parent representations to better handle mismatch between ontologies and observed EHR patterns. KAME \citep{ma2018kame} incorporates ontology knowledge beyond embedding learning into the prediction pipeline. To better exploit hierarchy, HAP \citep{zhang2020hierarchical} propagates information through top-down and bottom-up attention. Beyond pure hierarchies, G-BERT \citep{shang2019pre} combines ontology graphs with BERT-style encoders, ADORE \citep{cheong2023adaptive} leverages relational ontologies such as SNOMED to integrate heterogeneous code systems, and KAMPNet \citep{an2023kampnet} adopts graph contrastive learning for EHR representations.

More recent work augments structured EHRs with external signals such as clinical text and web knowledge. GCL~\citep{lu2021collaborative} and RAM-EHR~\citep{xu2024ram} incorporate unstructured text and retrieved medical knowledge, while retrieval-augmented frameworks such as KARE~\citep{jiang2024reasoning} and GraphCare~\citep{jiang2023graphcare} integrate multi-source knowledge for representation learning. LINKO~\cite{nayebi2025multi} leverages multiple ontologies with intra- and inter-ontology propagation via multi-level graph attention.


\section{Conclusion}
\vspace{-1mm}
We presented \model{}, a framework that integrates evidence-grounded KG construction, LLM-based semantic enrichment, and joint LLM-GNN co-learning for medical concept representation. \model{} builds a heterogeneous diagnosis--medication--procedure KG from EHR-derived evidence, enriches it with LLM-inferred relations and textual semantics, and learns unified concept embeddings through a LoRA-tuned LLM encoder and a heterogeneous GNN. Experiments on MIMIC-III and MIMIC-IV show that \model{} consistently improves sequential diagnosis prediction across multiple backbones and outperforms strong baselines, especially on rare labels and in limited-data settings. Additional analyses show \model{}'s strong clinical validity, robustness across diverse LLM encoders, and high inference efficiency.
\section{Limitations}

While \model{} amortizes LLM computation via a coverage-aware update schedule, joint KG--LLM training can still be resource-intensive, and scaling to larger vocabularies, longer contexts, or larger LLM backbones may increase computational cost.

\section{Ethical Considerations}

We comply with the ACL Ethics Policy throughout this study. All datasets used are publicly available and contain de-identified patient records, providing strong privacy protections. We do not send any patient-level data to external or public LLM services; all LLM interactions are restricted to general, concept-level information (e.g., medical code descriptions and relation prompts).

\section{Acknowledgments}
We thank Zhan Ye, MD, and Guangfan Zhang, MD, for their expert clinical validation of our methodology and results. This research was supported by National Science Foundation (Award No. 2531881).

\bibliography{ref}

\vspace{0.5cm}
\appendix

\section{Appendix}

\subsection{Existing Medical Ontologies and Coding Resources}
\label{app:ontologies}

This appendix summarizes widely used medical ontologies and coding resources and highlights why they are insufficient, on their own, for learning heterogeneous medical concept representations over diagnosis (\textit{dx}), medication (\textit{rx}), and procedure (\textit{px}) codes. While these resources provide valuable standardization, they often lack high-coverage, actionable cross-domain relations (e.g., \textit{dx-rx}, \textit{dx-px}, \textit{px-rx}) and/or do not reflect cohort- and workflow-specific dependencies observed in real-world EHR data.

\paragraph{ICD and CCS.}
ICD \cite{who_icd} is one of the most widely adopted clinical coding systems, and CCS \cite{hcup_ccs_2025} provides clinically meaningful groupings of ICD codes for downstream analysis. Both are primarily hierarchical: they organize diagnoses and procedures through parent-child structure (or category groupings in CCS), which is effective for standardization and coarse abstraction. However, they do not provide explicit semantics linking diagnoses, medications, and procedures, and therefore offer limited support for modeling cross-domain clinical dependencies needed for predictive tasks.

\paragraph{ATC.}
The Anatomical Therapeutic Chemical (ATC) classification system \cite{who_atc} organizes drug concepts into a hierarchical taxonomy based on therapeutic and chemical characteristics. This structure is useful for grouping medications and supporting within-drug semantic similarity. However, ATC does not explicitly encode care relationships that connect medications to diagnoses or procedures (e.g., treatment vs.\ contraindication links, peri-procedural medication dependencies), limiting its ability to capture heterogeneous clinical mechanisms across \textit{dx/rx/px} domains.

\paragraph{SNOMED CT.}
SNOMED CT \cite{stearns2001snomed} is a richer clinical ontology that supports description-logic–based concept definition and classification through a large hierarchy and a constrained set of definitional attributes governed by its concept model (MRCM). These attributes can include, for example, finding site, causative agent, method, and active ingredient, enabling structured definitional semantics and fine-grained clinical concept modeling. Nevertheless, SNOMED CT is not designed to systematically encode guideline-oriented, cross-domain care relationships—such as drug-disease treatment/contraindication or procedure-disease indication and workflow dependencies—at the breadth and granularity typically required for clinical prediction from EHRs.

\paragraph{UMLS.}
The UMLS Metathesaurus \cite{bodenreider2004unified} provides an integration layer across major biomedical vocabularies (e.g., SNOMED CT, RxNorm, ICD-10-CM, CPT/HCPCS) by aligning synonymous meanings via shared Concept Unique Identifiers (CUIs). This normalization capability is highly valuable for cross-system alignment and harmonization. However, UMLS largely functions as a mapping and aggregation resource rather than a comprehensive repository of actionable clinical relations. Although it contains some curated cross-type links (e.g., drug-disease associations via medication-focused sources such as MED-RT), it does not consistently provide broad, high-coverage procedure-disease indications or procedure-drug semantics at the granularity needed for predictive modeling.

\paragraph{Summary.}
In summary, commonly used resources provide strong standardization, hierarchical organization, and normalization, but they do not fully capture the heterogeneous, cross-domain dependencies that drive real-world care trajectories in EHRs. Many clinically important \textit{dx-rx}, \textit{px-dx}, and \textit{px-rx} relations remain missing or fragmented, motivating approaches that can induce a more complete relational structure grounded in empirical EHR evidence.  Table~\ref{tab:ontology_limits} provides a compact comparison of each resource's structure/goal, scope, cross-domain coverage, and key limitations.

\begin{table*}[t]
\centering
\footnotesize
\setlength{\tabcolsep}{3pt}
\renewcommand{\arraystretch}{1.15}
\begin{tabular}{p{2cm} p{3cm} p{2cm} p{1.8cm} p{6.5cm}}
\toprule
\textbf{Resource} & \textbf{Structure / Goal} & \textbf{Scope} & \textbf{Cross-Type} & \textbf{Key Limitations} \\
\midrule
ICD / CCS &
Coding taxonomy &
dx, px & No &
Primarily within-dx or within-px hierarchical categories with only parent-child relations; lacks explicit cross-domain relations. \\

ATC &
Drug classification taxonomy &
rx & No &
Drug taxonomy; primarily within-rx hierarchical categories with only parent-child relations; lacks explicit cross-domain relations. \\

SNOMED CT &
Definitional concept model &
dx, px, rx & Limited &
Rich definitional semantics (e.g., finding site, causative agent, method, active ingredient); limited coverage of within-domain relations (dx-due-to-dx); no coverage of cross-domain relations at predictive granularity. \\

UMLS &
Metathesaurus integration / mapping &
dx, rx, px & Limited &
Strong normalization/mapping; cross-type relations are incomplete/uneven (e.g., limited coverage for px-dx indications and px-rx semantics). \\
\bottomrule
\end{tabular}
\caption{Common medical resources (ICD \cite{who_icd}, CCS \cite{hcup_ccs_2025}, ATC \cite{who_atc}, SNOMED CT \cite{stearns2001snomed}, UMLS \cite{bodenreider2004unified}) and their limitations.}
\label{tab:ontology_limits}
\end{table*}

\subsection{Statistics of Final Evidence-Supported Code Pairs}
\label{app:pair_stats}

\begin{figure}[h!]
    \centering
    \includegraphics[width=0.5\textwidth]{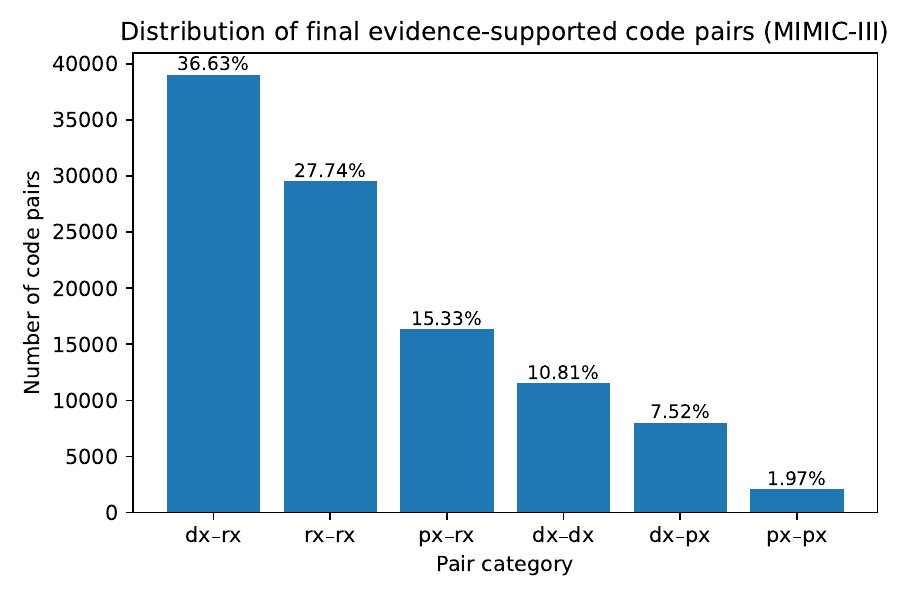}
    \caption{Distribution of final evidence-supported code pairs by category (MIMIC-III). Percentage values are shown on top of each bar.}
    \label{fig:pair_category_barplot_mimic3}
\end{figure}

\begin{figure}[h!]
    \centering
    \includegraphics[width=0.5\textwidth]{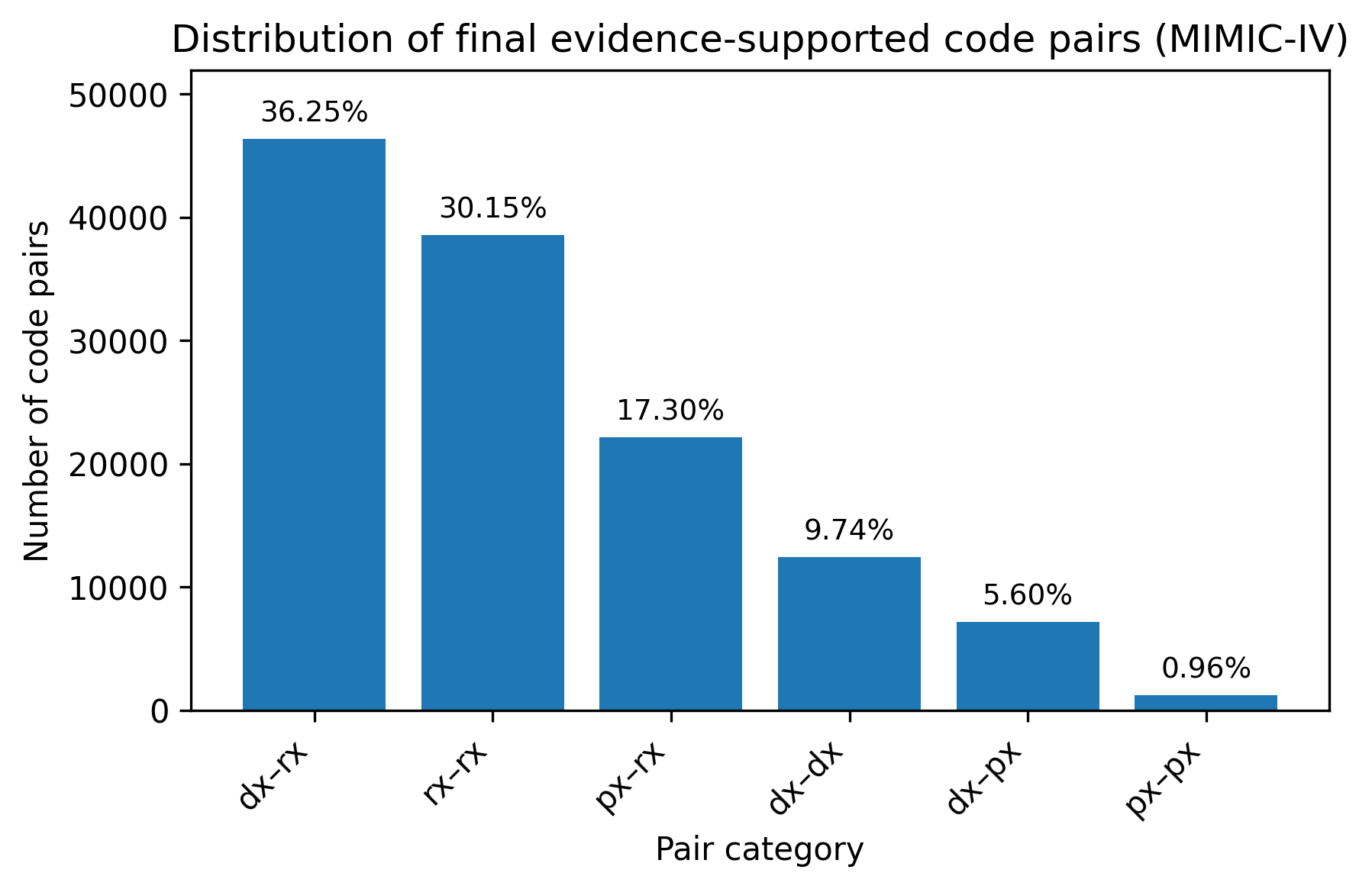}
    \caption{Distribution of final evidence-supported code pairs by category (MIMIC-IV). Percentage values are shown on top of each bar.}
    \label{fig:pair_category_barplot_mimic4}
\end{figure}

Figures~\ref{fig:pair_category_barplot_mimic3} and~\ref{fig:pair_category_barplot_mimic4} summarize the categorical distribution of the final set of statistically supported code-code pairs extracted from the MIMIC-III and MIMIC-IV cohorts. These pairs represent the output of the evidence extraction and filtering pipeline described in Section~\ref{sec:kg_construction}, and serve as input to the LLM-based relation inference stage.

\subsection{Clinical Relation Inventory for Code–Code Edges}
\label{app:ontology}

We develop a type-constrained relationship inventory comprising 28 clinically meaningful relationships that span all pairs of code categories (diagnosis, medication, procedure). Tables~\ref{tab:ontology_dx_dx}-\ref{tab:ontology_px_rx} enumerate the full relation sets and their definitions. These relations were selected to capture the major semantic patterns observed in clinical care, including etiologic links, disease progression, therapeutic and diagnostic intent, procedural workflow, pharmacologic interactions, safety considerations, and contextual co-occurrence patterns.

This relationship inventory is designed to be (i) clinically grounded, (ii) interpretable for downstream KG reasoning tasks, and (iii) sufficiently expressive to cover the majority of meaningful inter-code relationships in structured EHR data, while maintaining conservative abstention categories (\textit{no\_significant\_relation}, \textit{cannot\_decide}) for ambiguous or low-evidence cases.

\subsection{LLM-inferred relation distribution.}
\begin{figure*}[t]
  \centering
  \includegraphics[width=0.85\textwidth]{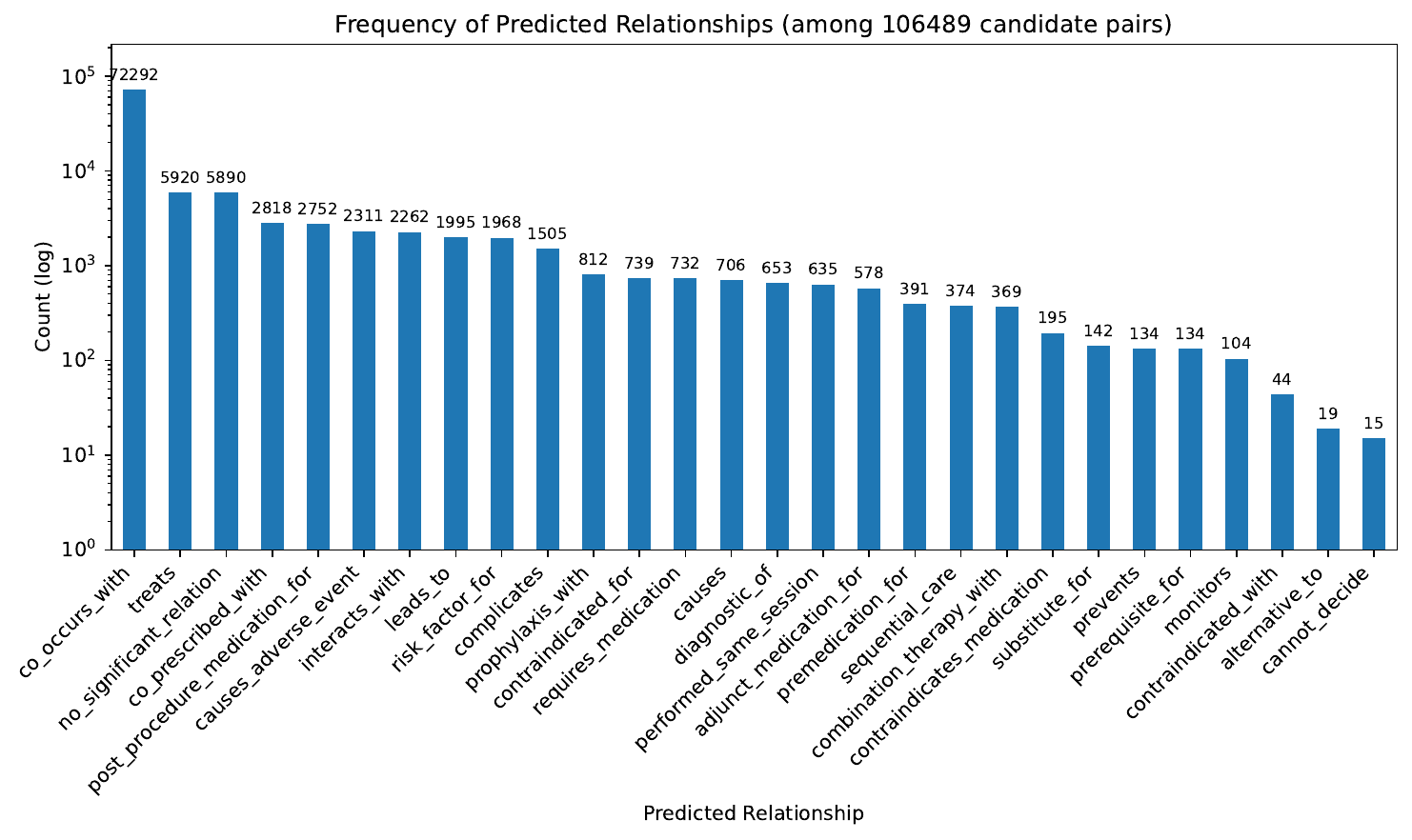}
  \caption{Frequency of LLM-predicted relation labels over evidence-filtered candidate code pairs in MIMIC-III. The y-axis is log-scaled to highlight the long-tailed distribution of relation types.}
  \label{fig:rel_dist_mimic3}
\end{figure*}

\begin{figure*}[t]
  \centering
  \includegraphics[width=0.85\textwidth]{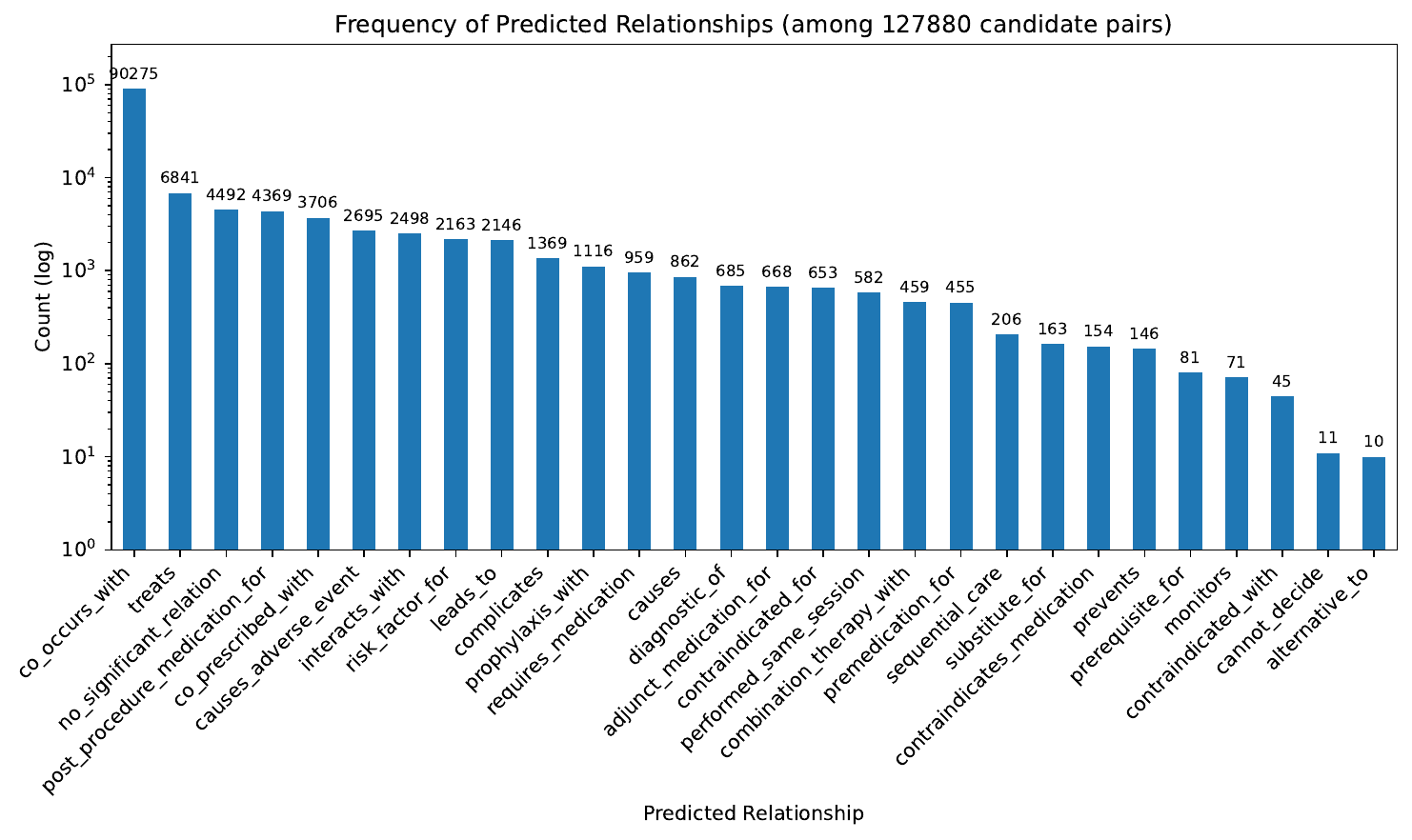}
  \caption{Frequency of LLM-predicted relation labels over evidence-filtered candidate code pairs in MIMIC-IV. The y-axis is log-scaled to highlight the long-tailed distribution of relation types.}
  \label{fig:rel_dist_mimic4}
\end{figure*}

Figures~\ref{fig:rel_dist_mimic3} and~\ref{fig:rel_dist_mimic4} summarize the distribution of relation labels predicted by the LLM over the evidence-filtered candidate code pairs, using the type-constrained relation inventory in our prompt. The resulting histogram is strongly long-tailed (shown on a log-scaled y-axis): a small number of generic relations account for most assignments, while many clinically specific relations occur much less frequently. This pattern is expected in real-world EHR data, where broad statistical association is common, but high-specificity semantics (e.g., contraindications, substitutions, or rare adverse interactions) require stronger or more distinctive evidence and therefore appear sparsely. Importantly, MIMIC-III and MIMIC-IV exhibit highly similar distributional shapes and rank orderings of the dominant relation types, suggesting that the prompting strategy yields stable and reproducible relation labeling across dataset versions. Any moderate shifts in mid-frequency labels are plausibly attributable to differences in cohort size, coding practices, and temporal coverage rather than prompt instability.

\subsection{Clinical Expert Audit of LLM-Inferred Edges}
\label{app:expert_audit}

To assess the clinical validity of LLM-inferred relations and quantify hallucination risk, we conducted a targeted expert audit of edges in the constructed knowledge graph. We sampled 50 edges from the final KG by selecting 5 edges from each of the 10 most frequent predicted relation types. The audit was performed under general clinical knowledge rather than patient-specific context.

Each sampled edge was independently reviewed by two frontline clinicians from our university health system. Reviewers rated the correctness of each edge on a 5-point scale:
\textit{1 = wrong},
\textit{2 = somewhat wrong},
\textit{3 = not sure / depends on context},
\textit{4 = somewhat correct}, and
\textit{5 = correct}.
For each edge, we computed the mean of the two clinician ratings, and we then report mean $\pm$ standard deviation across all audited edges and within each relation type.

Overall, the sampled edges received a mean rating of $4.84 \pm 0.29$, suggesting that the induced relations are largely clinically valid. As shown in Table~\ref{tab:expert_audit_edges}, all audited relation types achieved a mean score of at least 4.40, and several relation types (\textit{post\_procedure\_medication\_for}, \textit{risk\_factor\_for}, and \textit{treats}) achieved perfect mean scores in the audited sample. The lowest-scoring category was \textit{no\_significant\_relation} ($4.40 \pm 0.42$), which is unsurprising because abstention-style labels can be more context-sensitive than strongly expressed clinical relations.

These findings are consistent with the safeguards built into our KG induction pipeline. First, candidate code pairs are proposed only when supported by strong EHR association signals after statistical filtering. Second, the LLM is constrained to a predefined, type-specific relation inventory and is allowed to abstain using \textit{cannot\_decide} or \textit{no\_significant\_relation}. Third, we apply post-processing checks to remove invalid labels, directionality inconsistencies, and low-confidence outputs. While this audit is limited in scale, it provides an initial expert validation that the induced KG relations are clinically meaningful and that the evidence-grounded prompting strategy substantially mitigates unsupported edge generation.

\begin{table}[t]
\centering
\small
\begin{tabular}{lcc}
\toprule
\textbf{Predicted relationship} & \textbf{Mean} & \textbf{Std} \\
\midrule
causes\_adverse\_event & 4.70 & 0.27 \\
co\_occurs\_with & 4.80 & 0.45 \\
co\_prescribed\_with & 4.80 & 0.27 \\
complicates & 4.90 & 0.22 \\
interacts\_with & 4.90 & 0.22 \\
leads\_to & 4.90 & 0.22 \\
no\_significant\_relation & 4.40 & 0.42 \\
post\_procedure\_medication\_for & 5.00 & 0.00 \\
risk\_factor\_for & 5.00 & 0.00 \\
treats & 5.00 & 0.00 \\
\midrule
\textbf{Overall} & \textbf{4.84} & \textbf{0.29} \\
\bottomrule
\end{tabular}
\caption{Clinical expert audit of sampled LLM-inferred edges. We sampled 50 edges by selecting 5 edges from each of the 10 most frequent predicted relation types. Each edge was independently rated by two frontline clinicians on a 1-5 correctness scale, and the reported values are computed from the per-edge mean of the two ratings.}
\label{tab:expert_audit_edges}
\end{table}

\begin{table*}[t]
\centering
\small
\caption{Diagnosis-Diagnosis (DX-DX) relation types. Here, \textit{codeA} and \textit{codeB} are diagnosis codes.}
\label{tab:ontology_dx_dx}
\begin{tabular}{lp{0.68\textwidth}}
\toprule
\textbf{Relation} & \textbf{Description} \\
\midrule
\texttt{causes} &
codeA is an etiologic cause of codeB (direct causal or pathophysiologic link). \\[2pt]
\texttt{risk\_factor\_for} &
codeA increases the likelihood of codeB (epidemiologic association; not necessarily causal). \\[2pt]
\texttt{leads\_to} &
codeA typically precedes codeB as a downstream condition or stage (temporal progression without requiring strict causality). \\[2pt]
\texttt{complicates} &
codeA occurs as a complication during the course of codeB (arises secondary to codeB or its treatment). \\[2pt]
\texttt{co\_occurs\_with} &
codeA and codeB are frequently observed together without implied direction or clear causality (contextual association). \\[2pt]
\texttt{no\_significant\_relation} &
Clinical prior and available data indicate no clinically meaningful association between codeA and codeB. \\[2pt]
\texttt{cannot\_decide} &
Insufficient or conflicting evidence; the model should abstain from assigning a relation. \\
\bottomrule
\end{tabular}
\end{table*}

\begin{table*}[t]
\centering
\small
\caption{Medication-Diagnosis (RX-DX) relation types. \textit{codeA} is a medication and \textit{codeB} is a diagnosis.}
\label{tab:ontology_rx_dx}
\begin{tabular}{lp{0.68\textwidth}}
\toprule
\textbf{Relation} & \textbf{Description} \\
\midrule
\texttt{treats} &
codeA treats or manages codeB (therapeutic use). \\[2pt]
\texttt{prevents} &
codeA reduces risk or recurrence of codeB (prophylaxis). \\[2pt]
\texttt{causes\_adverse\_event} &
codeA may induce codeB as an adverse reaction. \\[2pt]
\texttt{contraindicated\_for} &
codeA should be avoided when codeB is present. \\[2pt]
\texttt{co\_occurs\_with} &
codeA and codeB often co-occur without a clear causal link. \\[2pt]
\texttt{no\_significant\_relation} &
Clinical prior and available data indicate no clinically meaningful association between codeA and codeB. \\[2pt]
\texttt{cannot\_decide} &
Insufficient or conflicting evidence; the model should abstain from assigning a relation. \\
\bottomrule
\end{tabular}
\end{table*}

\begin{table*}[t]
\centering
\small
\caption{Procedure-Diagnosis (PX-DX) relation types. \textit{codeA} is a procedure and \textit{codeB} is a diagnosis.}
\label{tab:ontology_px_dx}
\begin{tabular}{lp{0.68\textwidth}}
\toprule
\textbf{Relation} & \textbf{Description} \\
\midrule
\texttt{diagnostic\_of} &
codeA is performed to diagnose, confirm, or rule in/out codeB (diagnostic evaluation). \\[2pt]
\texttt{treats} &
codeA is a procedure used to treat, correct, or palliate codeB (therapeutic intervention). \\[2pt]
\texttt{monitors} &
codeA is performed to monitor, follow, or assess disease activity or treatment response for codeB. \\[2pt]
\texttt{contraindicated\_for} &
codeA should be avoided when codeB is present due to safety or unfavorable risk-benefit considerations. \\[2pt]
\texttt{co\_occurs\_with} &
codeA and codeB frequently appear together without a clear diagnostic, therapeutic, or causal link (contextual association). \\[2pt]
\texttt{no\_significant\_relation} &
Clinical prior and available data indicate no clinically meaningful association between codeA and codeB. \\[2pt]
\texttt{cannot\_decide} &
Insufficient or conflicting evidence; the model should abstain from assigning a relation. \\
\bottomrule
\end{tabular}
\end{table*}

\begin{table*}[t]
\centering
\small
\caption{Medication-Medication (RX-RX) relation types. \textit{codeA} and \textit{codeB} are medications.}
\label{tab:ontology_rx_rx}
\begin{tabular}{lp{0.68\textwidth}}
\toprule
\textbf{Relation} & \textbf{Description} \\
\midrule
\texttt{co\_prescribed\_with} &
codeA and codeB are intentionally prescribed together in practice. \\[2pt]
\texttt{contraindicated\_with} &
Concomitant use of codeA and codeB is contraindicated due to serious safety risk. \\[2pt]
\texttt{interacts\_with} &
codeA and codeB have a clinically meaningful drug-drug interaction (pharmacokinetic and/or pharmacodynamic) that may impact safety or efficacy. \\[2pt]
\texttt{substitute\_for} &
codeA is commonly used as a therapeutic alternative to codeB for similar indications (typically not co-prescribed). \\[2pt]
\texttt{combination\_therapy\_with} &
codeA and codeB are used together as an established combination therapy. \\[2pt]
\texttt{co\_occurs\_with} &
codeA and codeB frequently appear together in data without a known therapeutic relationship or interaction. \\[2pt]
\texttt{no\_significant\_relation} &
Clinical prior and available data indicate no clinically meaningful association between codeA and codeB. \\[2pt]
\texttt{cannot\_decide} &
Insufficient or conflicting evidence; the model should abstain from assigning a relation. \\
\bottomrule
\end{tabular}
\end{table*}

\begin{table*}[t]
\centering
\small
\caption{Procedure-Procedure (PX-PX) relation types. \textit{codeA} and \textit{codeB} are procedures.}
\label{tab:ontology_px_px}
\begin{tabular}{lp{0.68\textwidth}}
\toprule
\textbf{Relation} & \textbf{Description} \\
\midrule
\texttt{sequential\_care} &
codeA is typically followed by codeB as the next procedural step (ordered workflow; not necessarily causal). \\[2pt]
\texttt{prerequisite\_for} &
codeA is commonly required or preparatory for performing codeB (e.g., access, imaging guidance, setup). \\[2pt]
\texttt{alternative\_to} &
codeA and codeB are procedural alternatives for a similar clinical purpose (mutually substitutable options). \\[2pt]
\texttt{performed\_same\_session} &
codeA and codeB are commonly performed during the same procedural session or time block within an episode (intentional bundling). \\[2pt]
\texttt{co\_occurs\_with} &
codeA and codeB occur within the same episode without implied order or intentional bundling (contextual association). \\[2pt]
\texttt{no\_significant\_relation} &
Clinical prior and available data indicate no clinically meaningful association between codeA and codeB. \\[2pt]
\texttt{cannot\_decide} &
Insufficient evidence: no strong clinical prior and statistical signals are low-support, unstable, or conflicting; the model should abstain from assigning a relation. \\
\bottomrule
\end{tabular}
\end{table*}

\begin{table*}[t]
\centering
\small
\caption{Procedure-Medication (PX-RX) relation types. \textit{codeA} is a procedure and \textit{codeB} is a medication.}
\label{tab:ontology_px_rx}
\begin{tabular}{lp{0.68\textwidth}}
\toprule
\textbf{Relation} & \textbf{Description} \\
\midrule
\texttt{requires\_medication} &
codeA routinely requires administration of codeB as part of the procedure (e.g., sedation, anesthesia, anticoagulation). \\[2pt]
\texttt{premedication\_for} &
codeB is typically given before codeA to enable or optimize the procedure (e.g., anxiolysis, antibiotic prophylaxis). \\[2pt]
\texttt{post\_procedure\_medication\_for} &
codeB is commonly given after codeA for recovery, prophylaxis, or symptom control (e.g., analgesia, anticoagulation). \\[2pt]
\texttt{prophylaxis\_with} &
codeB is used to prevent complications associated with codeA (e.g., peri-procedural antibiotics, DVT prophylaxis). \\[2pt]
\texttt{adjunct\_medication\_for} &
codeB is an adjunct given with codeA to improve efficacy, safety, or tolerability (non-essential but commonly used). \\[2pt]
\texttt{contraindicates\_medication} &
When codeA is planned or present, codeB should be avoided due to safety or risk-benefit concerns. \\[2pt]
\texttt{co\_occurs\_with} &
codeA and codeB are observed together in data without a clear procedural rationale or causal link (contextual association). \\[2pt]
\texttt{no\_significant\_relation} &
Clinical prior and available data indicate no clinically meaningful association between codeA and codeB. \\[2pt]
\texttt{cannot\_decide} &
Insufficient or conflicting evidence; the model should abstain from assigning a relation. \\
\bottomrule
\end{tabular}
\end{table*}

\subsection{LLM Prompts for Node-Level Clinical Descriptions}
\label{app:node_prompt}

\begin{figure*}[t] 
\vspace*{-5cm}
\includegraphics[width=1.05\textwidth]{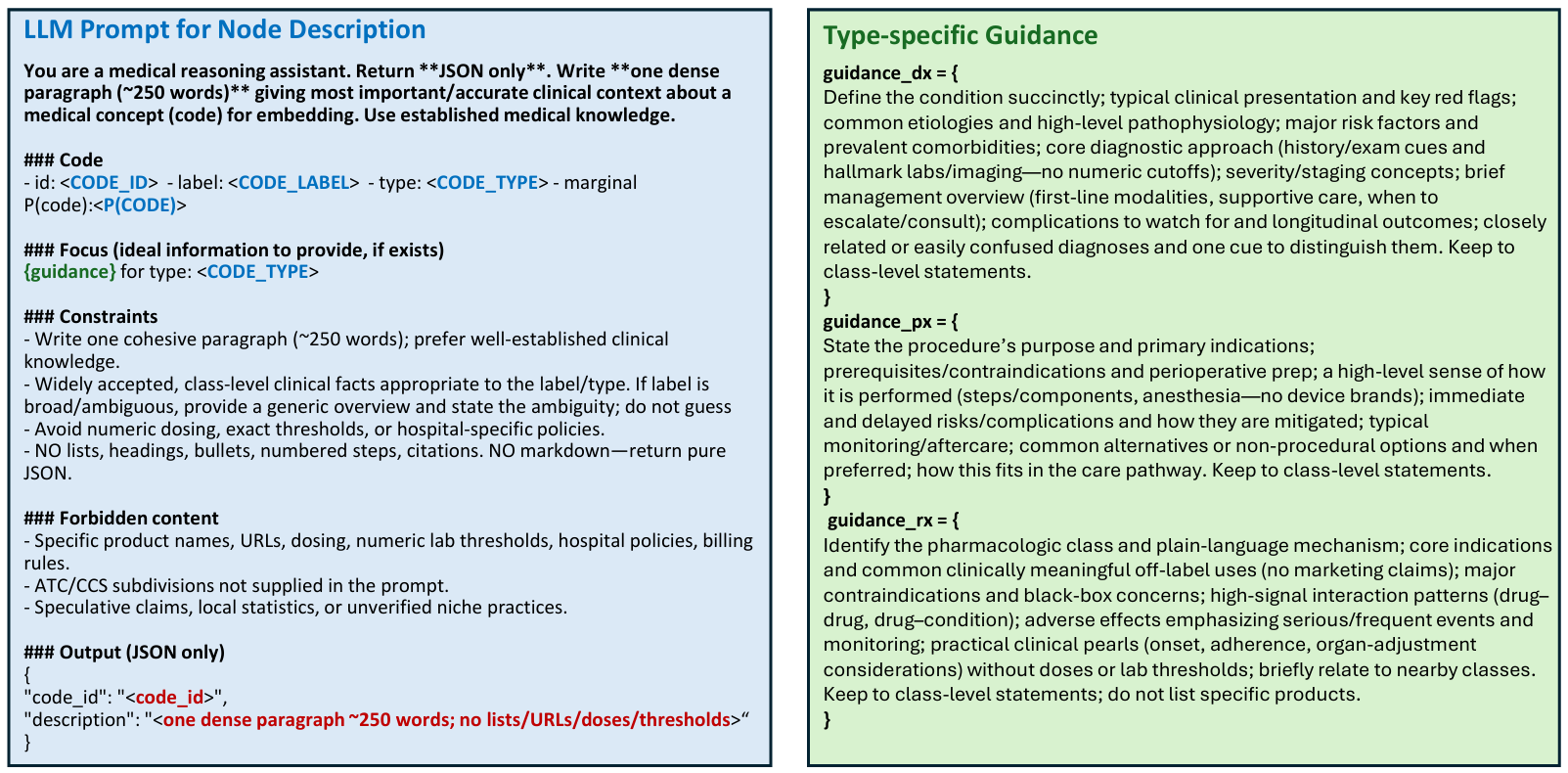} 
\caption{Prompt template used to generate LLM-based node-level clinical descriptions.} \label{fig:node_prompt} 
\end{figure*}

Figure~\ref{fig:node_prompt} illustrates the high-level structure of the prompt used to generate node-level clinical descriptions. Below, we provide the exact template used for all diagnosis (dx), procedure (px), and medication (rx) codes. Placeholders in angle brackets (e.g., \texttt{<CODE\_ID>}) are programmatically filled for each node.

\end{document}